# Phase Transitions and Backbones of the Asymmetric Traveling Salesman Problem

**Weixiong Zhang**      ZHANG@CSE.WUSTL.EDU
*Department of Computer Science and Engineering*
*Washington University in St. Louis*
*St. Louis, MO 63130, U.S.A.*

## Abstract

In recent years, there has been much interest in phase transitions of combinatorial problems. Phase transitions have been successfully used to analyze combinatorial optimization problems, characterize their typical-case features and locate the hardest problem instances. In this paper, we study phase transitions of the asymmetric Traveling Salesman Problem (ATSP), an NP-hard combinatorial optimization problem that has many real-world applications. Using random instances of up to 1,500 cities in which intercity distances are uniformly distributed, we empirically show that many properties of the problem, including the optimal tour cost and backbone size, experience sharp transitions as the precision of intercity distances increases across a critical value. Our experimental results on the costs of the ATSP tours and assignment problem agree with the theoretical result that the asymptotic cost of assignment problem is $\pi^2/6$ as the number of cities goes to infinity. In addition, we show that the average computational cost of the well-known branch-and-bound subtour elimination algorithm for the problem also exhibits a thrashing behavior, transitioning from easy to difficult as the distance precision increases. These results answer positively an open question regarding the existence of phase transitions in the ATSP, and provide guidance on how difficult ATSP problem instances should be generated.

## 1. Introduction and Overview

Phase transitions of combinatorial problems and thrashing behavior similar to phase transitions in combinatorial algorithms have drawn much attention in recent years (Gomes, Hogg, Walsh, & Zhang, 2001; Hogg, Huberman, & Williams, 1996; Martin, Monasson, & Zecchina, 2001). Having been extensively studied in the so-called spin glass theory (Mézard, Parsi, & Virasoro, 1987) in physics, phase transition refers to a phenomenon when some system properties change rapidly and dramatically when a control parameter undergoes a slight change around a critical value. Such transitions appear only in large systems. A larger system usually exhibits sharper and more abrupt phase transitions, leading to a phenomenon of crossover of the trajectories of phase transitions from systems of the same type but with different sizes.

A daily-life example of phase transitions is water changing from ice (solid phase) to water (liquid phase) to steam (gas phase) when temperature increases. It has been shown that many combinatorial decision problems have phase transitions, including Boolean satisfiability (Cheeseman, Kanefsky, & Taylor, 1991; Mitchell, Selman, & Levesque, 1992; Hogg, 1995; Selman & Kirkpatrick, 1996; Monasson, Zecchina, Kirkpatrick, Selman, & Troyansky, 1999), graph coloring (Cheeseman et al., 1991), and the symmetric Traveling Salesman Problem (deciding the existence of a complete tour of vising a given set of cities with a cost less than a specified value) (Cheeseman et al., 1991; Gent & Walsh, 1996a). Phase transitions can be used to characterize typical-case properties of difficult combinatorial problems (Gomes et al., 2001; Martin et al., 2001). The hardest problem instances of most decision problems appear very often at the points of phase transitions. Therefore, phase transitions have been used to help generate the hardest problem instances for testing and comparing algorithms for decision problems (Achlioptas, Gomes, Kautz, & Selman, 2000; Cheeseman et al., 1991; Mitchell et al., 1992).

Another important and useful concept for characterizing combinatorial problems is that of the backbone (Kirkpatrick & Toulouse, 1985; Monasson et al., 1999). A backbone variable refers to a variable that has a fixed value among all optimal solutions of a problem; and all such backbone variables are collectively referred to as the backbone of the problem. If a problem has a backbone variable, an algorithm will not find a solution to the problem until the backbone variable is set to its correct value. Therefore, the fraction of backbone variables, the percentage of variables being in the backbone, reflects the constrainedness of the problem and directly affects an algorithm searching for a solution. The larger a backbone, the more tightly constrained the problem becomes. As a result it is more likely for an algorithm to set a backbone variable to a wrong value, which may consequently require a large amount of computation to recover from such a mistake.

However, the research on the phase transitions and (particularly) backbones of *optimization problems* is limited, which is in sharp contrast with the numerous studies of the phase transitions and backbones of decision problems, represented by Boolean satisfiability (e.g., Cheeseman et al., 1991; Mitchell et al., 1992; Hogg, 1995; Selman & Kirkpatrick, 1996; Monasson et al., 1999). An early work on the symmetric TSP introduced the concept of backbones (Kirkpatrick & Toulouse, 1985). However, it has left the question whether there exists a phase transition of the TSP, the optimization version of the problem to be specific, open since 1985. One of the best (rigorous) phase-transition results was obtained on number partitioning (Borgs, Chayes, & Pittel, 2001), an optimization problem. However, the phase transition analyzed by Borgs, Chayes, & Pittel (2001), also experimentally by Gent & Walsh (1996b, 1998), is the existence of a perfect partition for a given set of integers, which is in essence a decision problem. In addition, Gent & Walsh (1996b, 1998) also studied the phase transitions of the size of optimal 2-way partition. The relationship between the phase transitions of satisfiability, a decision problem, and maximum satisfiability, an optimization problem, was studied by Zhang (2001). It was experimentally shown that the backbone of maximum Boolean satisfiability also exhibits phase transitions, emerging from nonexistence to almost full size abruptly and dramatically. In addition, the relationship between backbones and average-case algorithmic complexity has also been considered (Slaney & Walsh, 2001).

In this paper, we investigate the phase transitions of the asymmetric Traveling Salesman Problem. The Traveling Salesman Problem (TSP) (Gutin & Punnen, 2002; Lawler, Lenstra, Kan, & Shmoys, 1985) is an archetypical combinatorial optimization problem and one of the first NP-hard problems studied (Karp, 1972). Many concepts, such as backbone (Kirkpatrick & Toulouse, 1985), and general algorithms, such as linear programming (Dantzig, Fulkerson, & Johnson, 1959), branch-and-bound (Little, Murty, Sweeney, & Karel, 1963), local search (Lin & Kernighan, 1973) and simulated annealing (Kirkpatrick, Gelatt, & Vecchi, 1983) were first introduced and studied using the TSP. The problem is also very often a touchstone for combinatorial algorithms. Furthermore, the fact that many real-world problems, such as scheduling and routing, can be cast as TSPs has made the problem of practical importance. In this paper, we consider the asymmetric TSP (ATSP), where a distance from one city to another may not be necessarily the same as the distance on the reverse direction. The ATSP is more general and most ATSPs are more difficult to solve than their symmetric counterparts (Johnson, Gutin, McGeoch, Yeo, Zhang, & Zverovitch, 2002).

Using the general form of the problem, i.e., the ATSP, we provide a positive answer to the longstanding open question posted by Kirkpatrick & Toulouse (1985) regarding the existence of phase transitions in the problem, and disapprove the claim made by Kirkpatrick & Selman (1994) that the Traveling Salesman Problem does not have a clear phase transition.

Specifically, using uniformly random problem instances of up to 1,500 cities, we empirically reveal that the average optimal tour length, the accuracy of the most effective lower-bound function for the problem (the assignment problem, see Martello & Toth (1987)), and the backbone of the ATSP undergo sharp phase transitions. The control parameter is the precision of intercity distances which is typically represented by the maximum number of digits for the distances. Note that these results are algorithm independent and are properties of the problem. Furthermore, we show that the average computational cost of the well-known branch-and-bound subtour elimination algorithm (Balas & Toth, 1985; Bellmore & Malone, 1971; Smith, Srinivasan, & Thompson, 1977) for the ATSP exhibits a phase-transition or thrashing behavior in which the computational cost grows abruptly and dramatically from easy to difficult as the distance precision increases. Our results lead to a practical guidance on how to generate large, difficult random problem instances for the purpose of algorithm comparison.

It is worthwhile to mention that besides the results by Kirkpatrick & Toulouse (1985) there are three additional pieces of early empirical work related to the phase transitions of the Traveling Salesman Problem. The research by Zhang & Korf (1996) investigated the effects of two different distance distributions on the average complexity of the subtour elimination algorithm for the asymmetric TSP. The main result is that the average complexity of the algorithm is controlled by the number of distinct distances of a random asymmetric TSP. We will extend this result further in Section 6. However, we need to caution that these results are algorithm specific, which may not necessarily reflect intrinsic features of the underlying problem. The research on phase transitions by Cheeseman, Kanefsky, & Taylor (1991) studied the decision version of the symmetric TSP (Cheeseman, 1991). A more thorough investigation on this issue was also carried out (Gent & Walsh, 1996a). Specifically, Gent & Walsh (1996a) analyzed the probability that a tour whose length is less than a specific value exists for a given random symmetric euclidean TSP, showing that the probability has a one-to-zero phase transition as the length of the desired tour increases. Note that the phase-transition results by Cheeseman, Kanefsky, & Taylor (1991, 1996a) do not address the open question by Kirkpatrick & Toulouse (1985) which is about the optimization version of the problem. The experimental results by Gent & Walsh (1996a) also showed that the computational cost of a branch-and-bound algorithm, which unfortunately was not specified in the paper, exhibits an easy-hard-easy pattern.

The paper is organized as follows. In Section 2, we describe the ATSP and a related problem called assignment problem (AP). We then investigate the parameter that controls phase transitions in Section 3, and study various phase transitions of the ATSP in Section 4. In Section 5 we analyze asymptotic ATSP tour cost, AP cost and the precision of AP as a heuristic function for solving the ATSP as the number of cities grows to a large number. In Section 6, we describe the well-known subtour elimination algorithm for the ATSP, and analyze the thrashing behavior of this algorithm. We discuss related work in Section 7 and finally conclude in Section 8.

## 2. The Asymmetric Traveling Salesman Problem and Assignment Problem

Given $n$ cities and the distance or cost between each pair of cities, the *Traveling Salesman Problem* (TSP) is to find a minimum-cost tour that visits each city once and returns to the starting city (Gutin & Punnen, 2002; Lawler et al., 1985). When the distance from city $i$ to city $j$ is the same as the distance from $j$ to $i$, the problem is the symmetric TSP (STSP). If the distance from city $i$ to city $j$ is not necessarily equal to the reverse distance, the problem is the asymmetric TSP (ATSP). The ATSP is more difficult than the STSP, with respect to both optimization and approximation (Johnson et al., 2002). The TSPs are important NP-hard problems (Garey & Johnson, 1979; Karp, 1972) and have many practical applications. Many difficult combinatorial optimization problems, such as vehicle routing, workshop scheduling and computer wiring, can be formulated and solved as the TSPs (Gutin & Punnen, 2002; Lawler et al., 1985).

The ATSP can be formulated as an integer linear programming problem. Let $V$ be the set of $n$ cities, $A$ the set of all pairs of cities, and $D = (d_{ij})$ the distance or cost matrix specifying the distance of each pair of cities. The following integer linear programming formulation of the ATSP is well known:

$$ATSP(D) = \min \sum_{i,j} d_{ij} x_{ij}, \tag{1}$$

subject to

$$\sum_{i \in V} x_{ij} = 1, \quad \forall j \in V; \tag{2}$$

$$\sum_{j \in V} x_{ij} = 1, \quad \forall i \in V; \tag{3}$$

$$\sum_{i \in S} \sum_{j \in S} x_{ij} \leq |S| - 1, \quad \forall S \subset V, S \neq \emptyset; \tag{4}$$

$$x_{ij} \geq 0, \quad \forall i, j \in V \tag{5}$$

where variables $x_{ij}$ take values zero or one, and $x_{ij} = 1$ if and only if arc $(i, j)$ is in the optimal tour, for $i$ and $j$ in $V$. Constraints (2) and (3) restrict the in-degree and out-degree of each city to be one, respectively. Constraints (4) impose the *subtour elimination* constraints so that only complete tours are allowed.

The ATSP is closely related to the *assignment problem* (AP) (Martello & Toth, 1987), which is to assign to each city $i$ another city $j$, with the distance from $i$ to $j$ as the cost of the assignment, such that the total cost of all assignments is minimized. The AP is a relaxation of the ATSP in which the assignments need not form a complete tour. In other words, by removing the subtour elimination constraints (4) from the above representation, we have an integer linear programming formulation of the AP. Therefore, the AP cost is a lower bound on the ATSP tour cost. If the AP solution happens to be a complete tour, it is also a solution to the ATSP. While the ATSP is NP-hard, the AP can be solved in polynomial time, in $O(n^3)$ to be precise (Martello & Toth, 1987).

## 3. The Control Parameter

Consider two cases of the ATSP, one with all the intercity distances being the same and the other with every intercity distance being distinct. In the first case, every complete tour going through all $n$ cities is an optimal tour or a solution to the ATSP. There is no backbone variable since removing one edge from an optimal solution will not prevent finding another optimal solution. The ATSP in this case is easy; finding an optimal solution does not require any search at all. In addition, the cost of the optimal solution is also a constant, which is $n$ times of the intercity distance. In the second case where all distances are distinct, every complete tour covering all $n$ cities has a high probability to have a distinct cost. Therefore, an arc in the optimal solution is almost surely a backbone variable and removing it may destroy the optimal solution. In addition, it is expected to be difficult to find and verify such an optimal solution among a large number of suboptimal solutions in this case.

Therefore, there are significant differences between the above two extreme cases. One of the most important differences is the number of distinct distances in the distance matrix $D$. More precisely, many important characteristics of the random ATSP, including the size of backbone and complexity, are determined by the fraction of distinct distances among all distances. We denote the fraction of distinct distances in distance matrix $D$ as $\rho$. We are particularly interested in determining how $\rho$ affects the characteristics of the ATSP when it gradually increases from zero, when all distances are the same, to one, when all distances are distinct.

In practice, however, we do not directly control the number or the fraction of distinct distances in matrix $D$. Besides the actual structures of the "layouts" of the cities, the precision of the distances affects the number of distinct distances. The precision of a number is usually represented by the maximal number of digits allowed for the number. This is even more so when we use a digital computer to solve the ATSP, which typically has 32 bits (or 4 bytes) for integers or 64 bits (or 8 bytes) for double precision real numbers. As a result, the number of digits for distances is naturally a good choice for the control parameter.

The effect of a given number of digits on the fraction of distinct distances in distance matrix $D$ is relative to the problem size $n$. Consider a matrix $D$ with distances uniformly chosen from integers $\{0, 1, 2, \cdots, R-1\}$, where the range $R$ is determined by the number of digits $b$. For a fixed number of digits $b$, the fraction of distinct distances of a larger matrix $D$ is obviously smaller than that of a smaller $D$. Therefore, the control parameter for the fraction $\rho$ of distinct distances of $D$ must be a function of the number of digits $b$ and the number of cities $n$, which we denote as $\rho(n, b)$.

To find the control parameter, consider the number of distinct distances in $D$ for a given integer range $R$. The problem of finding the number of distinct distances is equivalent to the following bin-ball problem. We are given $M$ balls and $R$ bins, and asked to place the balls into the bins. Each ball is independently put into one of the bins with an equal probability. We are interested in the fraction of bins that are not empty after all the placements. Here, for asymmetric TSP $M = n^2 - n$ balls correspond to the total number of nondiagonal distances of matrix $D$, and $R$ bins represent the possible integers to choose from. Since each ball (distance) is thrown independently and uniformly into one of the $R$ bins (integers), the probability that one bin is not empty after throwing $M$ balls is $1 - (1 - 1/R)^M$. The expected number of occupied bins (distinct distances) is simply $R\left(1 - (1 - 1/R)^M\right)$. Thus, the expected fraction of distinct distances in matrix $D$ is

$$E[\rho(n, b)] = \frac{R\left(1 - (1 - 1/R)^M\right)}{M} \qquad (6)$$

Note that if $M$ or $n$ is fixed, $E[\rho(n, b)] \to 1$ as $R \to \infty$, since in this case the expectation of the number of distinct distances approaches $M$. On the other hand, when $R$ is fixed, $E[\rho(n, b)] \to 0$ when $M$ or $n$ goes to infinity, since all of a finite number of $R$ bins will be occupied by an infinite number of balls in this case.

ZHANGFollowing convention in practice, we use decimal values for distances. Thus $R = 10^b$, where $b$ is the number of digits for distances. It turns out that if we plot $\rho(n, b)$ against $b/\log_{10}(n)$, it will have relatively the same scale for different problem sizes $n$. This is shown in Figure 1(a). This means that the scaling function for the effective number of digits is $f(n) = \log_{10}^{-1}(n)$. Function $b/\log_{10}(n)$ is thus the *effective number of digits* that controls the fraction of distinct distances in matrix $D$, which we denote as $\beta(n, b)$. This also means that to have the same effective number of digits $\beta$ on two different problem sizes, say $n_1$ and $n_2$ with $n_1 < n_2$, the range $R$ should be different. On these two problems, $R$ needs to be $n_1^\beta$ and $n_2^\beta$, respectively, giving $n_1^\beta < n_2^\beta$.

We need to point out that the integer range $R$ can also be represented by a number in other bases, such as binary. Which base to use will not affect the results quantitatively, but introduces a constant factor to the results. In fact, since $b = \log_{10}(R)$, where $R$ is the range of integers to be chosen, $\beta(n, b) = b/\log_{10}(n) = \log_n(R)$, which is independent of the base of the values for intercity distances.

It is interesting to note that, controlled by the effective number of digits $b/\log_{10}(n)$, the fraction of distinct entities $\rho$ has a property similar to a phase transition, also shown in Figure 1(a). The larger the problem, the sharper the transition, and there exists a crossover point among the transitions of problems with different sizes. We may examine this phase-transition phenomenon more closely using finite-size scaling. Finite-size scaling (Barber, 1983; Wilson, 1979) is a method that has been successfully applied to phase transitions in similar systems of different sizes. Based on finite-size scaling, around a critical parameter (temperature) $T_c$, problems of different sizes tend to be indistinguishable except for a change of scale given by a power law in a characteristic length, which is typically in the form of $(T - T_c)n^{1/v}$, where $n$ is the problem size and $v$ the exponent of the rescaling factor. Therefore, finite-size scaling characterizes a phase transition precisely around the critical point $T_c$ of the control parameter as the problem scales to infinity. However, our analysis revealed that the scaling factor has a large exponent of 9 (Zhang, 2003), indicating that the phase transitions in Figure 1(a) does not exactly follow the power law finite-size scaling.

To find the correct rescaled control parameter, we reconsider (6). As $n \to \infty$ and distance range $R$ grows with problem size $n$, i.e., $R \to \infty$ as $n \to \infty$, we can rewrite (6) as

$$\lim_{n\to\infty, R\to\infty} E[\rho(n, b)] = \lim_{R\to\infty} \frac{R}{M}\left(1 - \left((1 - 1/R)^R\right)^{M/R}\right)$$
$$= \frac{R}{M}\left(1 - e^{-M/R}\right), \quad (7)$$

where the second equation follows $\lim_{R\to\infty}(1 - 1/R)^R = e^{-1}$. Since our underlying control parameter is the number of digits, $b = \log_{10}(R)$, we take $x = \log_{10}(R/M)$. Asymptotically as $n \to \infty$, $M \simeq n^2$, which leads to $x = \log_{10}(R) - 2\log_{10}(n) = (\beta - 2)\log_{10}(n)$. Using $x$, we rewrite (7) as

$$\lim_{n\to\infty, R\to\infty} E[\rho(n, b)] = 10^x \left(1 - e^{-10^{-x}}\right). \quad (8)$$

The rescaled control parameter as $n \to \infty$ for the expected number of distinct distances in $D$ is $(\beta - 2)\log_{10}(n)$. Therefore, the critical point is 2 and the rescaling factor is $\log_{10}(n)$. The rescaled phase transition is shown in Figure 1(b), which plots $\rho(n, (\beta - 2)\log_{10}(n))$.

Note that the number of digits used for intercity distances is nothing but a measurement of the precision of the distances. The larger the number of digits, the higher the precision becomes. This

476

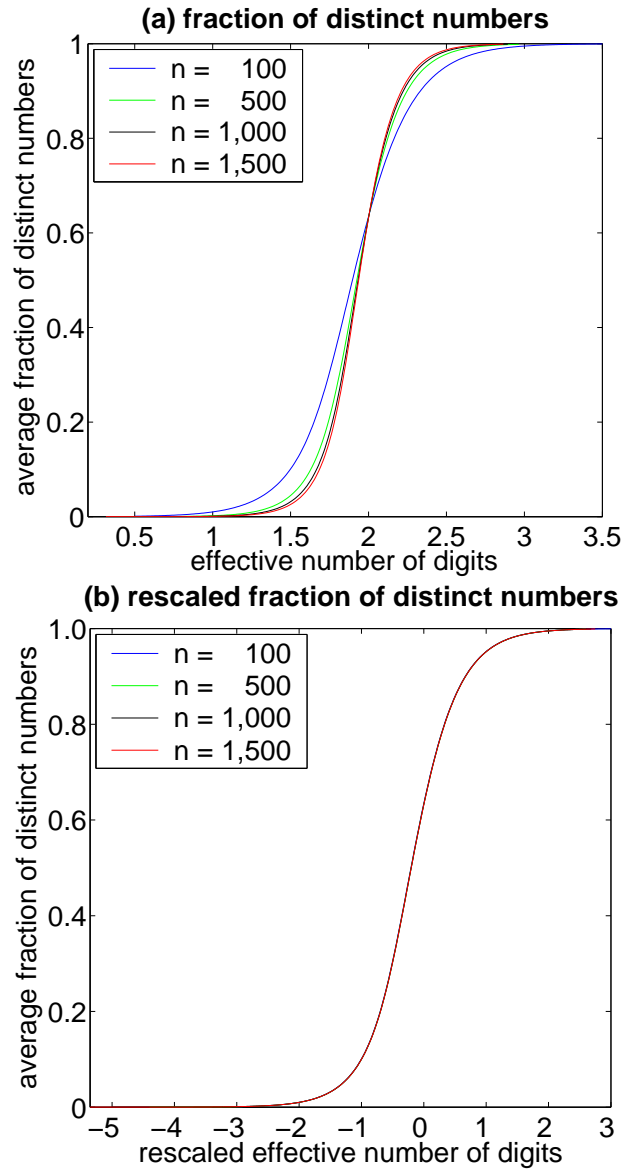

Figure 1: (a) Average fraction of distinct distances in matrix $D$, $\rho(n,b)$, controled by the effective number of digits, $\beta = b \log_{10}^{-1}(n)$, for $n = 100, 500, 1000$ and $1500$. (b) Average $\rho(n,b)$ after finite-size scaling, with scaling factor $(\beta - \beta_c) \log_{10}(n)$, where $\beta_c = 2$.

agrees with the common practice of using more effective digits to gain precision. Therefore, the control parameter is in turn determined by the precision of intercity distances.

Finally, it is important to note that even though the discussion in this section focused on asymmetric cost matrices and the ATSP, the arguments apply to symmetric distance matrices and the symmetric TSP as well. That is, with $M$ revised to $(n^2 - n)/2$, asymptotically as $R$ and $n$ goes to infinity, $\log_{10}(M) \simeq 2\log_{10}(n)$, so that $(\beta - 2)\log_{10}(n)$ is also a rescaled control parameter for the number of distinct distances in symmetric cost matrices.

## 4. Phase Transitions

With the control parameter, the effective number of digits $\beta(n, b)$ for intercity distances, identified, we are now in a position to investigate possible phase transitions in the ATSP and the related assignment problem.

To set forth to investigate these phase transitions, we generated and studied uniformly random problem instances with 100-, 200-, 300- upto 1,000-cities and 1,500-cities. Although we have results from 100-, 200-, 300-, up to 1,000-city as well as 1,500-city problems, to make the result figures readable, we only use the data from 100-, 500-, 1,000- and 1,500-city problems to report the results. For the problem instances considered, intercity distances are independently and uniformly chosen from $\{0, 1, 2, \cdots, R - 1\}$ for a given range $R$, which is controlled by the number of digits $b$. We varied $b$ from 1.0 to 6.0 for instances with up to 1,000-cities and from 1.0 to 6.5 for instances with 1,500-cities. The digits are incremented by 0.1, i.e., we used $b = 1.0, 1.1, 1.2, \cdots$.

### 4.1 Phase Transitions in the ATSP

We are particularly interested in possible phase transitions in the ATSP cost, phase transitions of backbones and phase transitions of the numbers of ATSP tours. The results on backbone can shed some light on the intrinsic tightness of the constraints among the cities as the precision of distance measurement changes.

#### 4.1.1 THE ATSP TOUR COST

There is a phase transition in the ATSP tour cost, $ATSP(D)$, under the control parameter $\beta$, the effective number of digits for intercity distances. Figure 2(a) shows the results on 100-, 500-, 1,000- and 1,500-city ATSP instances, averaged over 10,000 instances for each data point. The reported tour costs are obtained by dividing the integer ATSP tour costs by $n \times (R - 1)$, where $n$ is the number of cities and $R$ the range of intercity costs. Equivalently, an intercity distance was virtually converted to a real value in $[0, 1]$.

As shown, the ATSP tour cost increases abruptly and dramatically as the effective number of digits increases, exhibiting phase transitions. The transitions become sharper as the problem becomes larger, and there exist crossover points among curves from different problem sizes. By finite-size scaling, we further determine the critical value of the control parameter at which the phase transitions occur. Following the discussion in Section 3, the scaling factor has a form of $(\beta - \beta_c)\log_{10}(n)$. Our numerical result indicated that $\beta_c = 1.02 \pm 0.007$. We thus use $\beta_c = 1$ to show the result in Figure 2(b). It is worthwhile to mention that the AP cost follows almost the same phase-transition pattern of the ATSP tour cost in Figure 2.

#### 4.1.2 BACKBONE AND NUMBER OF OPTIMAL SOLUTIONS

We now turn to the backbone of the ATSP, which is the percentage of directed arcs that appear in all optimal solutions. The backbone also exhibits phase transitions as the effective number of digits for distances increases. The result is included in Figure 3(a), where each data point is averaged over 10,000 problem instances. The rescaled result is shown in Figure 3(b), where the critical point $\beta_c = 1$. Interestingly, the phase-transition pattern of the backbone follows a similar trend as that of the fraction of distinct entities in the distance matrix, shown in Figure 1. In addition, the phase-transition patterns of tour costs and backbones are similar, which will be discussed in Section 4.3.

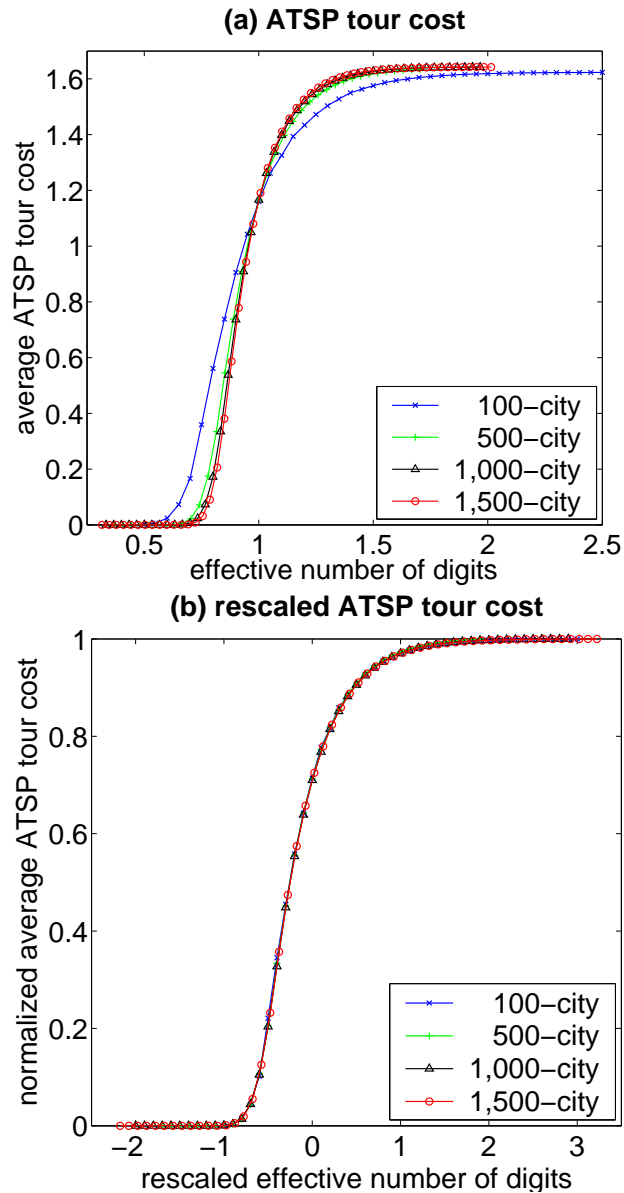

Figure 2: (a) Average optimal ATSP tour cost. (b) Scaled and normalized average optimal tour cost, with rescaling factor $(\beta - \beta_c) \log_{10}(n)$ and $\beta_c = 1$.

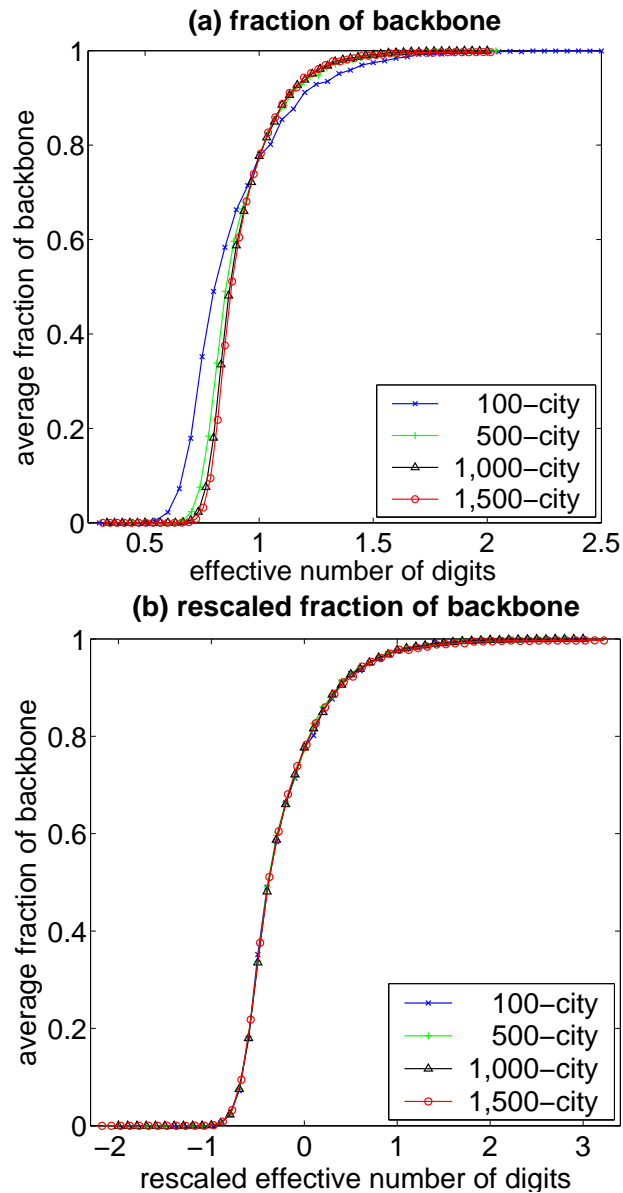

Figure 3: (a) Average fraction of backbone variables. (b) Rescaled average backbone fraction, with rescaling factor $(\beta - \beta_c) \log_{10}(n)$ and $\beta_c = 1$.

The fraction of backbone variables is related to the number of optimal solutions of a problem. We thus examined the total number of optimal solutions of the ATSP. This was done on small ATSPs, from 10 cities to 150 cities, as finding *all* optimal solutions on larger problems is computationally too expensive. The results are averaged over 100 trials for each data point. As shown in Figure 4, where the vertical axes are in logarithmic scale, the number of optimal tours also undergoes a phase

480

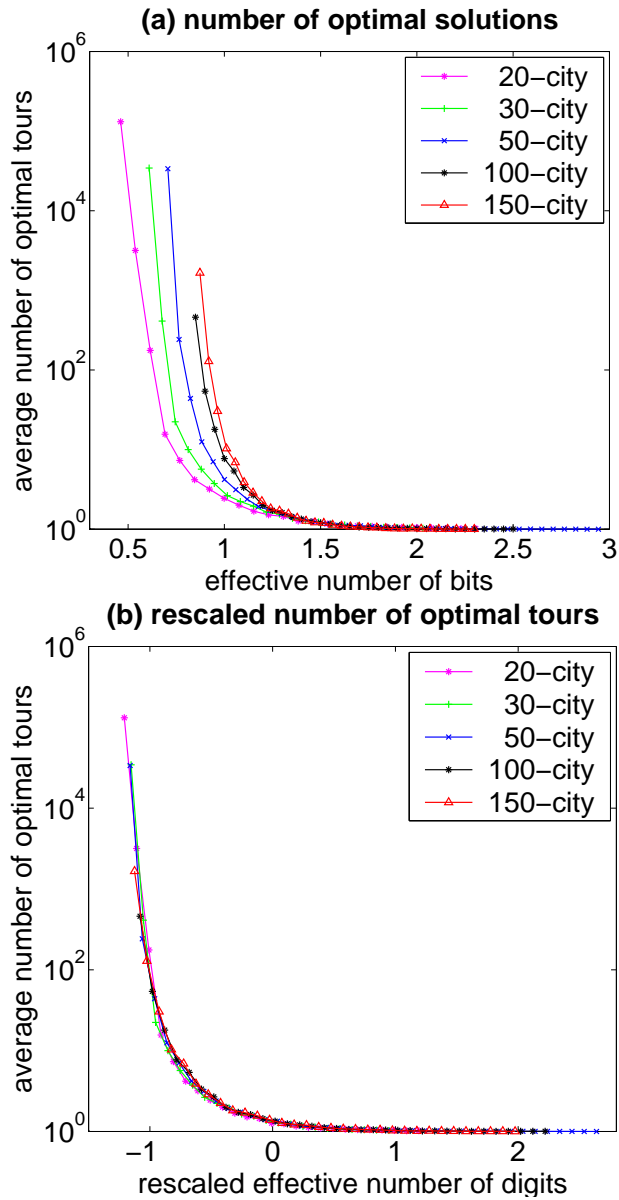

Figure 4: (a) Average number of optimal ATSP tours. (b) Rescaled average number of optimal ATSP tours, with rescaling factor $(\beta - \beta_c) \log_{10}(n)$ and $\beta_c = 1.39 \pm 0.008$.

transition, from exponential to a constant, as the number of digits increases. Note that when the number of digits is small, it is very costly to find all optimal solutions, even on these relatively small problems.

The fraction of backbone variables captures in essence the tightness of the constraints among the cities. As more intercity distances become distinct, the number of tours of distinct lengths increases. Consequently, the number of optimal solutions decreases and the fraction of backbone variables grows inversely. When more arcs are part of the backbone, optimal solutions become more restricted. As a result, the number of optimal solutions decreases. As the fraction of backbone variables increases and approaches one, the number of optimal solutions decreases and becomes one as well, which typically makes the problem of finding an optimal solution more difficult.

### 4.1.3 EXISTENCE OF HAMILTONIAN CIRCUIT WITH ZERO-COST EDGES

When the precision of intercity distances is low, it is likely that the ATSP tour, a complete tour of minimal cost among all complete tours, has a cost of zero, meaning that there exists an Hamiltonian circuit consisting of zero-cost arcs. It is a decision problem to determine if an Hamiltonian circuit exists in a given ATSP. We examined this decision problem using the same set of 10,000 problem instances used in Figures 2 and 3. The result is shown in Figure 5. Notice that although it follows the same rescaling formula of $(\beta - \beta_c) \log_{10}(n)$, the critical point of the transition, $\beta_0 = 0.865$, is different from the critical point of $\beta_c = 1$ for the phase transitions of backbones and ATSP tour cost, as shown in Figures 2 and 3.

## 4.2 Quality of the AP Lower-bound Function

The existence of Hamiltonian circuits of zero-cost arcs also indicates that when the number of digits for intercity distances is very small, for example, less than 1.9 (or $R \approx 80$) for $n = 1,500$, both the AP and ATSP costs are zero, meaning that these two costs are the same as well. It is useful to know how likely the AP cost is equal to the ATSP tour cost; the answers to this issue constitutes the first step to the elucidation of the accuracy of the AP lower-bound cost function.

Given a random distance matrix $D$, how likely is it that an AP cost will be the same as the ATSP tour cost as the effective number of digits $\beta$ increases? We answer this question by examining the probability that an AP cost $AP(D)$ is equal to the corresponding ATSP cost $ATSP(D)$ as $\beta$ increases. Figure 6(a) shows the results on 100-, 500-, 1,000- and 1,500-city ATSP instances, averaged over the same set of 10,000 instances for Figure 2 for each data point. As shown in the figure, the probability that $AP(D) = ATSP(D)$ also experiences abrupt and dramatic phase transitions. Figure 6(b) shows the phase transitions after finite-size scaling, with critical point $\beta_c = 1.17 \pm 0.005$.

The results in Figure 6 also imply that the quality of the AP lower-bound function degrades as the distance precision increases. The degradation should also follow a phase-transition process. This is verified by Figure 7, using the data from the same set of problem instances. Note that the critical point of the phase transition for the accuracy of AP is $\beta_c = 0.97$, which is different from the critical point $\beta_c = 1.17$ for the phase transition of the probability that $AP(D) = ATSP(D)$.

## 4.3 How Many Phase Transitions?

So far, we have seen many phase transitions on different features of the ATSP and its related assignment problem. Qualitatively, all these phase transitions follow the same transition pattern, meaning that they can all be captured by the same finite-size rescaling formula of $(\beta - \beta_c) \log_{10}(n)$, where $\beta_c$ is a critical point depending on the particular feature of interest.

It is interesting to note that the critical points for the phase transitions of the ATSP tour costs and fractions of backbone variables are all at $\beta_c = 1$. A close examination also indicates that these two phase transitions follow almost the same phase transition, as shown in Figure 8, where the rescaled curves for the ATSP tour cost and the fraction of backbone variables are drawn from 1,500-city ATSP, averaged over 10,000 problem instances.

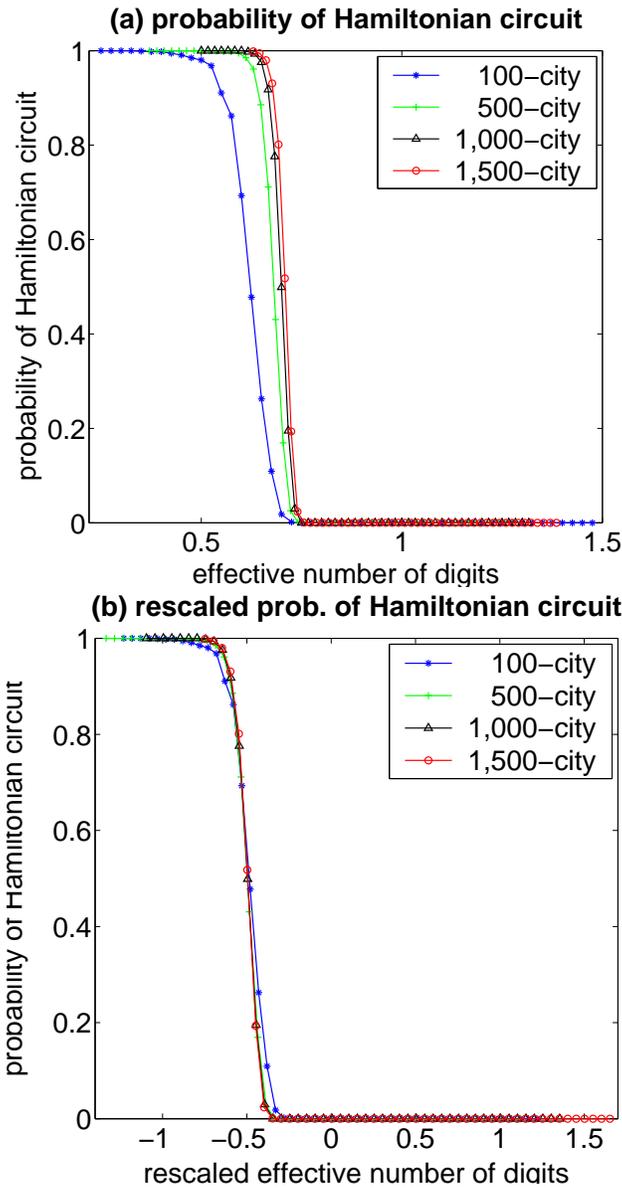

Figure 5: (a) Probability of the existence of Hamiltonian circuits with zero cost arcs. (b) Rescaled probability of zero-cost Hamiltonian circuits, with rescaling factor $(\beta - \beta_c) \log_{10}(n)$ and $\beta_c = 0.865$.

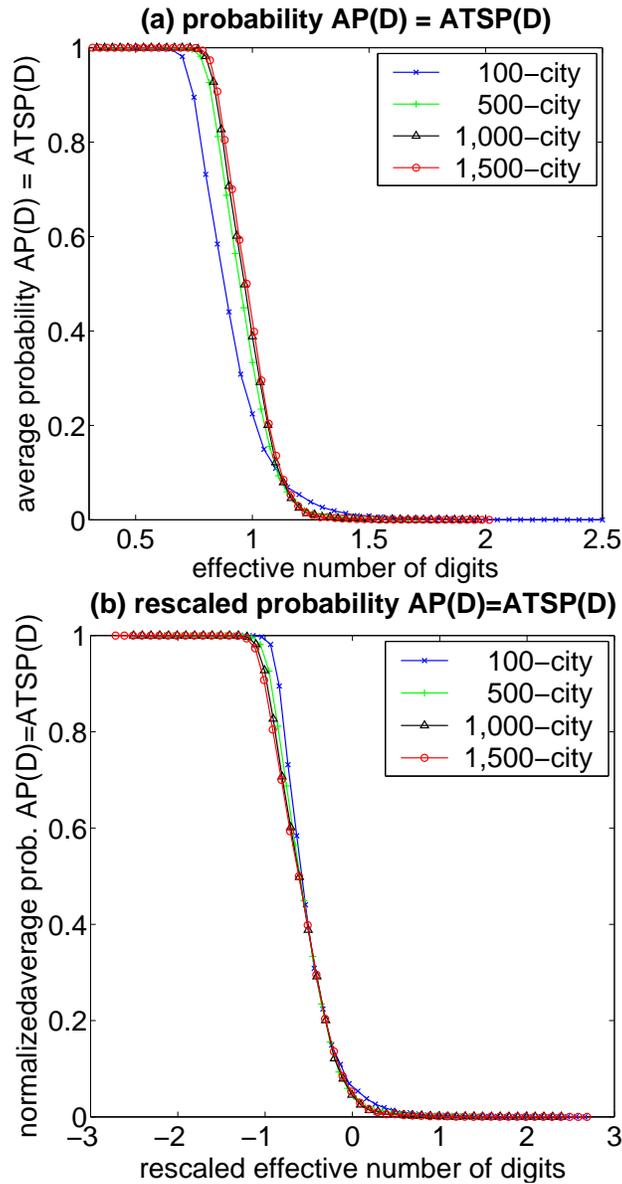

Figure 6: (a) Average probability that $AP(D) = ATSP(D)$. (b) Average probability after finite-size scaling, with rescaling factor $(\beta - \beta_c)\log_{10}(n)$ and $\beta_c = 1.17 \pm 0.005$.

Except the close similarity of the phase transitions of the ATSP tour cost and the fraction of backbone variables, the other phase transitions all have different critical points, indicating that they undergo the same type of phase transitions at different ranges.

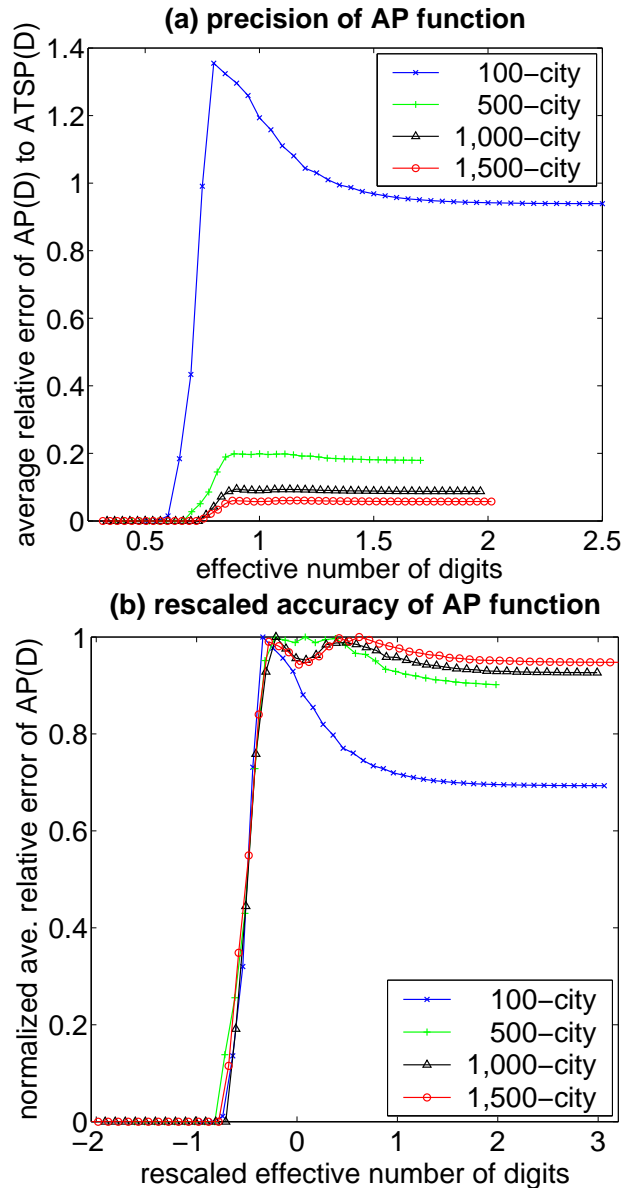

Figure 7: (a) Average accuracy of AP lower-bound function, measured by the error of AP cost relative to ATSP cost. (b) normalized and rescaled average accuracy, with rescaling factor $(\beta - \beta_c) \log_{10}(n)$ and $\beta_c = 0.97$.

## 5. Asymptotic ATSP Tour Length and AP Precision

As a by-product of the phase-transition results, we now provide numerical values of the ATSP cost, the AP cost and its accuracy, asymptotically when the number of cities grows. We attempt to

485

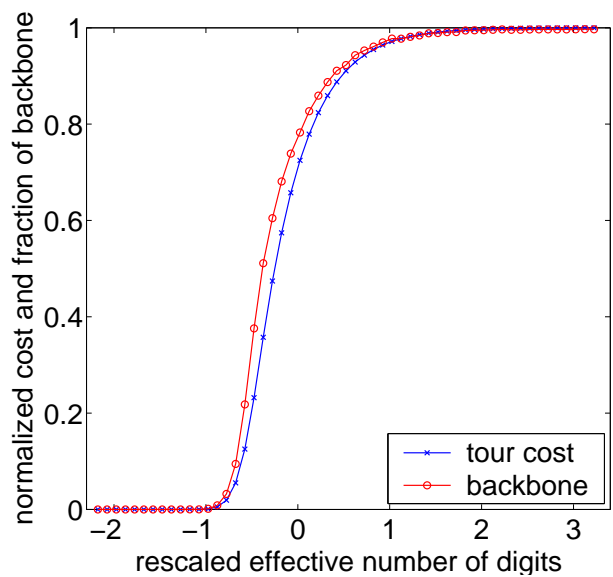

Figure 8: Simultanous examination of the phase transitions of backbone and ATSP tour cost on 1,500-city problems, all rescaled with $(\beta - 1) \log_{10}(n)$.

extend the previous theoretical results on the AP cost, which was shown to asymptotically approach $\pi^2/6$ (Aldous, 2001; Mézard & Parsi, 1987), and the observations that the relative error of the AP lower bounds decreases as the problem size increases (Balas & Toth, 1985; Smith et al., 1977).

Not every real number can be represented in a digital computer. Thus, it is infeasible to directly examine a theoretical result on reals using a digital computer. For our purpose, on the other hand, the phase-transition results indicate that when the precision of the intercity distances is high enough, all the quantities of the ATSP we have examined, including the ATSP cost, the AP cost and its precision as a lower-bound cost function, as well as the backbone, are relatively stable, in the sense that they do not change significantly even when the precision of intercity distances increases further. Therefore, it is sufficient to use a high distance precision to experimentally analyze the asymptotic properties of the ATSP cost and other related quantities.

We need to be cautious in selecting the number of digits for intercity distances. As we discussed in Section 3, the same number of digits for intercity distances gives rise to different effective numbers of digits on problems of different sizes. Furthermore, the phase transition results in Section 4 indicate that the effective numbers of digits must be scaled properly in order to have the same effect on problems of different sizes when we investigate an asymptotic feature.

Therefore, in our experiments, we fixed the scaled effective number of digits for intercity distances, $(\beta - \beta_c) \log_{10}(n)$, to a constant. Based on our phase-transition results, especially that on the control parameter in Figure 1, we chose to take $(\beta - 2) \log_{10}(n)$ a constant of 2.1, for two reasons. First, $(\beta - 2) \log_{10}(n) = 2.1$ is sufficiently large so that almost all distances are distinct, regardless of problem size, and the quantities we will examine will not change substantially after the finite-size scaling. Secondly, $(\beta - 2) \log_{10}(n) = 2.1$ is relatively small so that we can experiment on problems of large sizes. To save memory as much as possible, the intercity distances are integers of 4 bytes in our implementation of the subtour elimination algorithm. Thus the number of digits must be less than 9.4 without causing an overflow in the worst case. Using $(\beta - 2) \log_{10}(n) = 2.1$, we can go up to roughly 3,000-city ATSPs.

| $n$ | digits | AP cost | ATSP cost | relative AP error (%) |
|---:|---:|---:|---:|---:|
| 200 | 6.7021 | 1.63533 ± 0.00254 | 1.64302 ± 0.00254 | 0.46817 ± 0.00970 |
| 400 | 7.3041 | 1.63942 ± 0.00180 | 1.64311 ± 0.00180 | 0.22485 ± 0.00468 |
| 600 | 7.6563 | 1.64072 ± 0.00146 | 1.64314 ± 0.00145 | 0.14765 ± 0.00317 |
| 800 | 7.9062 | 1.64227 ± 0.00125 | 1.64407 ± 0.00125 | 0.10904 ± 0.00237 |
| 1,000 | 8.1000 | 1.64297 ± 0.00114 | 1.64441 ± 0.00114 | 0.08754 ± 0.00191 |
| 1,200 | 8.2584 | 1.64284 ± 0.00104 | 1.64402 ± 0.00105 | 0.07187 ± 0.00158 |
| 1,400 | 8.3923 | 1.64313 ± 0.00096 | 1.64413 ± 0.00096 | 0.06148 ± 0.00139 |
| 1,600 | 8.5082 | 1.64319 ± 0.00090 | 1.64405 ± 0.00090 | 0.05276 ± 0.00117 |
| 2,000 | 8.7021 | 1.64382 ± 0.00082 | 1.64451 ± 0.00082 | 0.04231 ± 0.00095 |
| 2,200 | 8.7848 | 1.64372 ± 0.00077 | 1.64434 ± 0.00077 | 0.03813 ± 0.00085 |
| 2,400 | 8.8604 | 1.64360 ± 0.00074 | 1.64417 ± 0.00073 | 0.03477 ± 0.00079 |
| 2,600 | 8.9299 | 1.64429 ± 0.00071 | 1.64481 ± 0.00071 | 0.03234 ± 0.00074 |
| 2,800 | 8.9943 | 1.64382 ± 0.00068 | 1.64430 ± 0.00068 | 0.02966 ± 0.00068 |
| 3,000 | 9.0542 | 1.64421 ± 0.00065 | 1.64463 ± 0.00065 | 0.02548 ± 0.00061 |

Table 1: Numerical results on AP cost, the ATSP cost and AP error relative to the ATSP cost, in percent. The cost matrices are uniformly random. Each data point is averaged over 10,000 problem instances. In the table, $n$ is the number of cities, digits is the number of digits for intercity distances, and all numerical error bounds represent 95 percent confidence intervals.

Table 1 shows the experimental results, with up to 3,000 cities, on the average AP cost, the ATSP tour cost, and accuracy of the AP cost function in the error of AP cost relative to the ATSP cost. The results are averaged over 10,000 instances for each problem size. Based on the results, the AP cost approaches to 1.6442 and the ATSP cost to 1.6446. Note that the experimental AP cost of 1.6442 is very close to the theoretical asymptotic AP cost of $\pi^2/6 \approx 1.6449$ (Aldous, 2001; Mézard & Parsi, 1987). In addition, the accuracy of AP function indeed improves as the problem size increases, reduced to about 0.02548% for 3,000-city problem instances. This result supports the previous observations (Balas & Toth, 1985; Smith et al., 1977).

## 6. Thrashing Behavior of Subtour Elimination

All the phase-transition results discussed in the previous section indicate that the ATSP becomes more constrained and difficult as the distance precision becomes higher. In this section, we study how a well-known algorithm for the ATSP, branch-and-bound subtour elimination (Balas & Toth, 1985; Bellmore & Malone, 1971; Smith et al., 1977), behaves. We separate this issue from the phase transition phenomena studied before because what we will consider in this section is the behavior of a particular algorithm, which may not be necessarily a feature of the underlying problem. Nevertheless, this is still an issue of its own interest because this algorithm is the oldest and is still among the best known methods for the ATSP, and we hope that a better understanding of an efficient algorithm for the ATSP can shed light on the typical case computational complexity of the problem.

### 6.1 Branch-and-bound Subtour Elimination

The branch-and-bound (BnB) subtour algorithm elimination (Balas & Toth, 1985; Bellmore & Malone, 1971; Smith et al., 1977) solves an ATSP in a state-space search (Pearl, 1984; Zhang, 1999) and uses the assignment problem (AP) as a lower-bound cost function. The BnB search takes the original ATSP as the root of the state space and repeats the following two steps. First, it solves the AP of the current problem. If the AP solution is not a complete tour, it decomposes it into subproblems by subtour elimination that breaks a subtour by excluding some arcs from a selected subtour. As a subproblem is more constrained than its parent problem, the AP cost to the subproblem must be as much as that of the parent. This means that the AP cost function is monotonically nondecreasing. While solving the AP requires $O(n^3)$ computation in general, the AP to a child node can be incrementally solved in $O(n^2)$ time based on the solution to the AP of its parent.

There are many heuristics for selecting a subtour to eliminate (Balas & Toth, 1985), and we use the Carpaneto-Toth scheme (Carpaneto & Toth, 1980), or the CT scheme for short, in our algorithm. One important feature of the CT scheme is that it generates no duplicate subproblem so that the overall search space is a tree. One example of this scheme is shown in Figure 9. The AP solution to the original ATSP contains two subtours that are in the root of the tree of the figure. The subtour $2-3-2$ is chosen to be eliminated, since it is shorter than the other subtour. We have two ways to break the selected subtour, i.e., excluding directed arc $(2,3)$ or $(3,2)$. Assume that we first exclude $(2,3)$ and then $(3,2)$, generating two subproblems, nodes $A$ and $B$ in Figure 9. When generating the second subproblem $B$, we deliberately include $(2,3)$ in its solution. By including the arc that was excluded in the previous subproblem $A$, we force to exclude in the current subproblem $B$ all the solutions to the original problem that will appear in $A$, and therefore form a partition of the solution space using $A$ and $B$. In general, let $\mathcal{E}$ be the excluded arc set, and $\mathcal{I}$ be the included arc set of the problem to be decomposed. Assume that there are $t$ arcs of the selected subtour, $\{e_1, e_2, \cdots, e_t\}$, that are not in $\mathcal{I}$. The CT scheme decomposes the problem into $t$ child subproblems, with the $k$-th one having excluded arc set $\mathcal{E}_k$ and included arc set $\mathcal{I}_k$, such that

$$\left. \begin{array}{ll} \mathcal{E}_k & = \mathcal{E} \cup \{e_k\}, \\ \mathcal{I}_k & = \mathcal{I} \cup \{e_1, \cdots, e_{k-1}\}, \end{array} \right\} k = 1, 2, \cdots, t. \qquad (9)$$

Since $e_k$ is an excluded arc of the $k$-th subproblem, $e_k \in \mathcal{E}_k$, and it is an included arc of the $k+1$-st subproblem, $e_k \in \mathcal{I}_{k+1}$, a complete tour obtained from the $k$-th subproblem does not contain arc $e_k$, while a tour obtained from the $k+1$-st subproblem must have arc $e_k$. Thus a tour from the $k$-th subproblem cannot be generated from the $k+1$-st subproblem, and vice versa. In summary, the state space of the ATSP under BnB using the CT subtour elimination scheme can be represented by a tree without duplicate nodes.

In the next step, the algorithm selects as the current problem a new subproblem from the set of *active subproblems*, which are the ones that have been generated but not yet expanded. This process continues until there is no unexpanded problem, or all unexpanded problems have costs greater than or equal to the cost of the best complete tour found so far.

Thanks to its linear-space requirement, we use depth-first branch-and-bound (DFBnB) in our algorithm. DFBnB explores active subproblems in a depth-first order. It uses an upper bound $\alpha$ on the optimal cost, whose initial value can be infinity or the cost of an approximate solution, such as one obtained by Karp's patching algorithm (Karp, 1979; Karp & Steele, 1985), which repeatedly patches two smallest subtours into a big one until a complete tour forms. Starting at the root node, DFBnB selects a recently generated node $x$ to examine next. If the AP solution of $x$ is a complete tour, then $x$ is a leaf node of the search tree. If the cost of a leaf node is less than the current upper bound $\alpha$, $\alpha$ is revised to the cost of $x$. If $x$'s AP solution is not a complete tour and its cost is greater than or equal to $\alpha$, $x$ is pruned, because node costs are monotonic so that no descendant of $x$ will have a cost smaller than $x$'s cost. Otherwise, $x$ is expanded, generating all its child nodes. To find an optimal goal node quickly, the children of $x$ should be searched in an increasing order of their costs. In other words we use node ordering to reduce the number of nodes explored. To speed up the process of reaching a better, possibly optimal, solution, we also apply Karp's patching algorithm to the best child node of the current node.

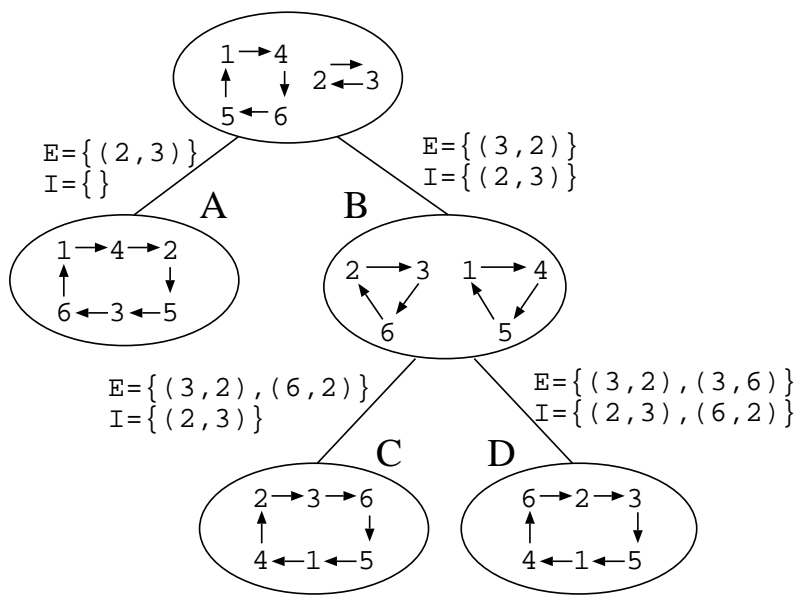

Figure 9: DFBnB subtour elimination on the ATSP.

Our algorithm is in principle the same as the algorithm by Carpaneto, Dell'Amico & Toth (1995), which is probably the best known complete algorithm for the ATSP. The main difference between the two is that, due to a consideration on space requirement, we use depth-first search while Carpaneto, Dell'Amico & Toth (1995) used best-first search.

### 6.2 Thrashing Behavior

The average computational complexity of the BnB subtour elimination algorithm is determined by two factors, the problem size, or the number of cities, and the number of digits used for intercity distances. Figure 10 illustrates this average complexity, measured by the number of calls to the AP function, in a logarithmic scale. The result is averaged over the same 10,000 problem instances for each data point as used for the phase transitions studied in Section 4. Note that the number of AP calls increases exponentially from small problems to large ones when they are generated using the same number of digits for distances.

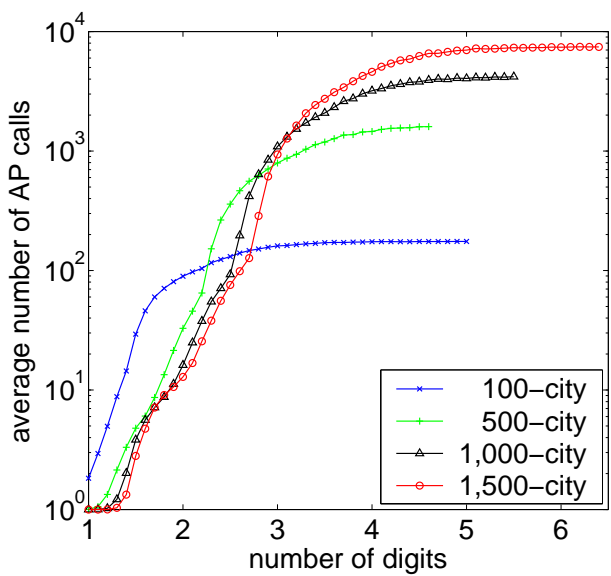

Figure 10: Average computational complexity of the BnB subtour elimination algorithm.

To characterize the thrashing behavior of the algorithm, we normalize the result in Figure 10 in such a way that for a given problem size, the minimal and maximal AP calls among all problem instances of the same size are mapped to zero and one, respectively, and the other AP calls are proportionally adjusted to a ratio between 0 and 1. This allows us to compare the results from different problem sizes in one figure. We also normalize the number of digits for distances by problem size. The curves in Figure 11(a) follow a pattern similar to that of the phase transitions in Section 4. The complexity of the subtour elimination algorithm increases with the effective number of digits, and exhibits a thrashing behavior similar to phase transitions. Indeed, we can use finite-size scaling to capture the behavior as the problem size grows, as illustrated in Figure 11(b).

The results in Figure 11 and the phase-transition results of Section 4 indicate that the complexity of the subtour elimination algorithm goes hand-in-hand with the accuracy of the AP function and the constrainedness of the problem, which is determined by the portion of distinct entities of distance matrix, which is in turn controlled by the precision of distances.

Similar results have been reported by Zhang & Korf (1996), where the effects of two different distance distributions on the average complexity of the subtour elimination algorithm were analyzed to conclude that the determinant of the average complexity is the number of distinct distances of a problem. The results of this section extend that by Zhang & Korf (1996) to different sizes of problems and by applying finite-size scaling to capture the thrashing behavior as problem size increases.

We need to contrast the experimental result in this section with the theoretical result on the NP-completeness of the TSP of intercity distances 0 and 1. It has been known that the degenerated TSP with distances 0 and 1 is still NP-complete (Papadimitriou & Yannakakis, 1993). On the other hand, our experimental results showed that when intercity distances are small, relative to the problem size, the ATSP is easy on average. Based on our experimental result, a large portion of the problem instances with small intercity distances can be solved by the assignment problem or Karp's patching algorithm with no branch-and-bound search required. This discrepancy indicates that the worst case of the problem is rare and most likely pathological.

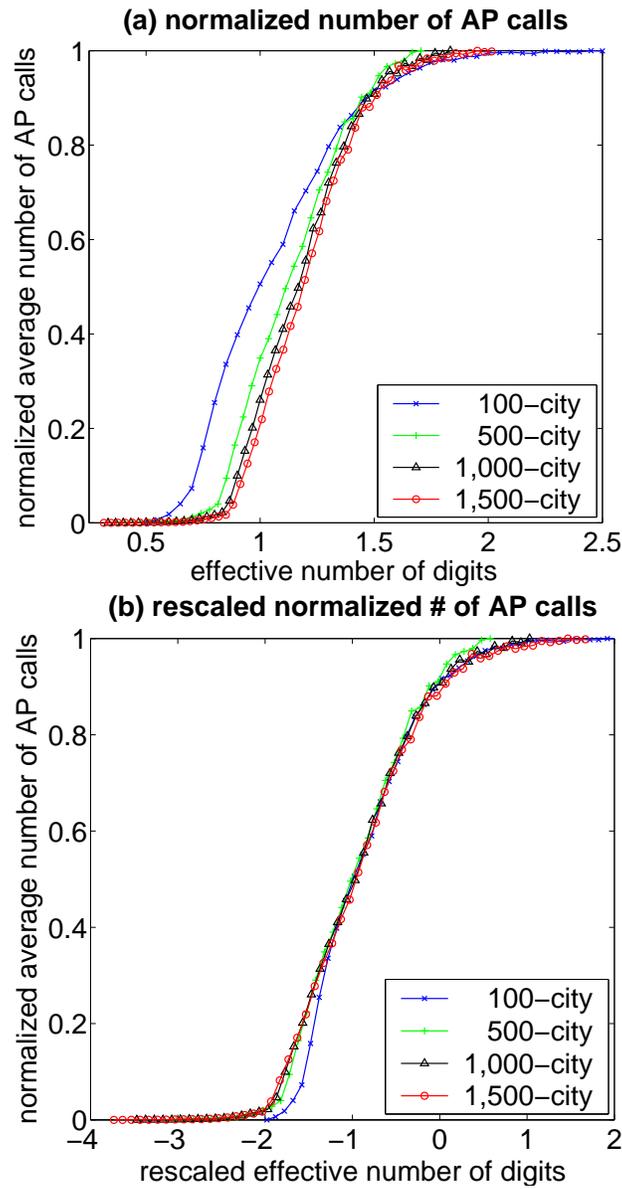

Figure 11: (a) Normalized average number of AP calls of DFBnB subtour elimination. (b) Scaled average number of AP calls, with $(\beta - \beta_c)\log_{10}(n)$, where $\beta_0 = 1.49 \pm 0.025$.

## 7. Related Work and Discussions

Two lines of previous work have directly influenced and inspired this research. The first line of related work was on the expected complexity of tree search, which shed light to the BnB subtour elimination algorithm described in Section 6.1 as it solves the ATSP in a tree search. The analysis was carried out on an abstract random tree model called *incremental tree* $T$ (Karp & Pearl, 1983; McDiarmid, 1990; McDiarmid & Provan, 1991; Zhang & Korf, 1995; Zhang, 1999). The internal nodes of $T$ has variable number of children and edges in $T$ are assigned finite and nonnegative random values. The cost of a node in $T$ is the sum of the edge costs along the path from the root to that node. An optimal goal node is a node of minimum cost at a fixed depth $d$. The overall goal is to find an optimal goal node.

There exist phase transitions in the cost of the optimal goal node and the complexity to the problem of finding an optimal goal in $T$. The control parameter is the expected number of child nodes of a common parent node which have the same cost as the parent. The cost of an optimal goal node almost surely undergoes a phase transition from a linear function of depth $d$ to a constant when the expected same-cost children of a node increases beyond one. Meanwhile, best-first search and depth-first branch-and-bound also exhibit a phase-transition behavior, i.e., their expected complexity changes dramatically from exponential to polynomial in $d$ as the expected same-cost children of a node is reduced to below one. Note that following the result by Dechter & Pearl (1985), best-first search is optimal for searching this random tree among all algorithms using the same cost function, in terms of number of node expansions, up to tie breaking. Thus, the above results also give the expected complexity of the problem of searching an incremental tree.

The second line of related research was on characterizing the the assignment problem (AP) lower-bound cost function and its relationship with the ATSP, which has been a research interest for a long time (Aldous, 2001; Coppersmith & Sorkin, 1999; Frieze, Karp, & Reed, 1992; Frieze & Sorkin, 2001; Karp, 1987; Karp & Steele, 1985; Mézard & Parsi, 1987; Walkup, 1979). The first surprising result (Walkup, 1979) is that the expected AP cost approaches a constant as the number of cities $n$ goes to infinity if the entries of distance matrix $D$ are independent and uniform over reals $[0, 1]$. This constant has been the subject of a long history of pursuit. It has been shown rigorously, based on rigorous replica method from statistical physics (Mézard et al., 1987), that the optimal cost of random assignment approaches asymptotically to $\pi^2/6$ (Aldous, 2001), which is approximately 1.64493. Our results in Section 5 show that the AP and the ATSP costs approach 1.64421 and 1.64463, respectively, which agree with the theoretical results on the AP cost.

More importantly, the relationship between the AP cost and the ATSP cost has remarkably different characteristics under different distance distributions. On one extreme, the AP cost is the same as the ATSP cost with a high probability, while on the other extreme, it can differ from the ATSP cost, with a high probability, by a function of problem size $n$. Let $AP(D)$ be the AP cost and $ATSP(D)$ the ATSP cost under a distance matrix $D$. If the expected number of zeros in a row of $D$ approaches infinity when $n \to \infty$, then $AP(D) = ATSP(D)$ with a probability tending to one (Frieze et al., 1992). However, if the entities of $D$ are uniform over the integers $[0, 1, \cdots, \lfloor c_n n \rfloor]$, then $AP(D) = ATSP(D)$ with a probability going to zero, where $c_n$ grows to infinity with $n$ (Frieze et al., 1992). Indeed, when the entities of $D$ are uniform over $[0, 1]$, $E(ATSP(D) - AP(D)) \geq c_0/n$, where $c_0$ is a positive constant (Frieze & Sorkin, 2001).

These previous results indicate that the quality of the AP function varies significantly, depending on the underlying distance distribution. Precisely, the difference between the AP cost and the ATSP cost has two phases, controlled by the number of zero distances in the distance matrix $D$. In one phase, the difference is zero with high probability, while in the other phase, the expectation of the difference is a function of the problem size $n$. Our experimental results in Section 4 adds to this analysis the existence of a phase transition between these two phases.

The two-phase result on the accuracy of the AP cost function is also in principle consistent with the phase-transition result of incremental random trees. The root of the search tree has a cost equal to the AP cost $AP(D)$ to the problem and an optimal goal node has the ATSP tour cost $ATSP(D)$. If we subtract the AP cost to the root from every node in the ATSP search tree, the root node has cost zero and an optimal goal node has cost equal to $ATSP(D) - AP(D)$. When there are a large number of zero distances in $D$, there will be a large number of same-cost children, and the AP cost of a child node in a search tree is more likely to be the same as the AP cost of its parent, since AP will tend to use the zero distances. Therefore, it is expected that more nodes in the search tree will have more than one child node having the same cost as their parents.

In addition to the phase transitions of combinatorial problems mentioned in Section 1, there are other related previous results. Results on scaling of search cost against constrainedness in symmetric TSP were considered by Gent, MacIntyre, Prosser, & T. Walsh (1997). Phase transitions in Hamiltonian circuit was studied by Frank, Gent, & Walsh (1998). It was also shown that it is hard to generate difficult Hamiltonian cycle problem instances (Vandegriend & Culberson, 1998). In addition, the concept of backbones has been studied in many problems under different names. For examples, unary prime implicate refers to such a variable that must be set to a fixed value for an instance of Boolean satisfiability (Parkes, 1997); a frozen development describes a pair of nodes that must share the same colors in a graph coloring problem (Culberson & Gent, 2001).

## 8. Conclusions

Our main contributions of this research are twofold. First, we answered positively the long-standing question if the Traveling Salesman Problem (TSP) has phase transitions (Kirkpatrick & Toulouse, 1985) and disapproved the belief that the problem does not have a phase transition (Kirkpatrick & Selman, 1994). We studied this issue on a more general, optimization version of the problem, the asymmetric TSP (ATSP). We empirically showed, using random problem instances with distances from a uniform distribution, that many important properties, including the ATSP tour cost and the fraction of backbone variables, have two characteristically different values, and the transitions between them are rather abrupt and dramatic, displaying a phase-transition phenomenon. The control parameter of the phase transitions is the effective number of digits representing the intercity distances or the precision of distance measure.

Second, our results provide a practical guidance on how to generate difficult random ATSP problem instances and which random instances should be used for comparing the asymptotic performance of ATSP algorithms. A current common practice in comparing algorithms when using a random ensemble is to generate problem instances of different sizes with a fixed distance precision. Our phase transition results indicate that the correct way is to use instances of different sizes that have the same or similar features such as the same fraction of backbone variables. It is also important to point out that the locations of hardest, albeit random, problem instances typically depend on distance distribution used. In the case of uniform distribution, this requires increasing the precision of intercity distances as the problem size grows.

It is important to note that the exact locations of various phase transitions presented here remain to be mathematically determined, using methods probably from statistical physics (Martin et al., 2001; Mézard et al., 1987).

We like to conclude by pointing out that the phase transition results in this paper is general. The argument on the control parameter in Section 3 is general and applicable to the *symmetric* TSP (STSP). Our unpublished data also showed phase transitions in the STSP. The results on the ATSP with uniformly distributed distances should hold for other types of intercity distances. This is in part supported by our previous investigation where intercity distances were chosen from a log-normal distribution (Zhang & Korf, 1996). Finally, we believe that phase transitions will persist on structured TSPs as long as intercity distances are independently drawn from a common distribution. Such TSPs include those proposed and studied by Cirasella et al., (2001) and Johnson et al., (2002), for examples, problem instances with constraints of triangle inequalities, instances converted from particular applications such as disk drive optimization, jobshop scheduling, coin collecting optimization, etc.

## Acknowledgments

This research was supported in part by NSF grants IIS-0196057 and ITR/EIA-0113618, and in part by DARPA Cooperative Agreements F30602-00-2-0531 and F33615-01-C-1897. Thanks to Sharlee Climer for the joint work on an algorithm for finding backbone using limit crossing (Climer & Zhang, 2002) and for a critical reading of a draft. Thanks also to Scott Kirkpatrick and David Johnson for comments on a draft. Special thanks to Allon Percus and Sergey Knysh for very constructive comments and suggestions, especially on the finite-size scaling, which significantly improved the paper. Thanks also go to the anonymous reviewers for excellent comments. Some early results were presented at *NSF/IPAM Workshop on Phase Transitions and Algorithmic Complexity*, June 3-5, 2002, and at *the 18-th International Joint Conference on Artificial Intelligence* (IJCAI-03), Acapulco, Mexico, Aug. 9-15, 2003 (Zhang, 2003).